\documentclass[letterpaper, 10 pt, journal, twoside]{ieeetran}

\IEEEoverridecommandlockouts                              

\usepackage{graphicx}
\usepackage{setspace}
\usepackage{acronym}
\usepackage{amsfonts}			
\usepackage{times}		
\usepackage{color,soul}
\usepackage{amssymb}
\usepackage{amsmath}
\usepackage{gensymb}
\usepackage{tabularx} 
\usepackage{multirow}
\usepackage{algorithm}
\usepackage{algorithmic}
\usepackage{booktabs}
\usepackage{physics}
\usepackage[T1]{fontenc}

\DeclareMathOperator*{\argmin}{arg\,min}
\usepackage{float}
\usepackage{mwe}
\newcolumntype{L}{>{\centering\arraybackslash}m{3cm}}
\usepackage[thinc]{esdiff}
\usepackage[table]{xcolor}
\usepackage{hyperref}
\hypersetup{
    colorlinks=true,
    citecolor=black,
    linkcolor=black,
    filecolor=black,      
    urlcolor=black,
    }
\usepackage{url}
\usepackage{array,ragged2e}
\newcolumntype{P}[1]{>{\RaggedRight\arraybackslash}p{#1}}

\newcolumntype{M}[1]{>{\centering\arraybackslash}m{#1}}

\newcommand*{\thead}[1]{\multicolumn{1}{c}{\bfseries #1}}
\usepackage[font=small,skip=2pt]{caption}
\usepackage{textcomp, gensymb}
\usepackage{dblfloatfix}
\usepackage{xcolor}
\usepackage{etoolbox}
\newbool{showcolor}
\setbool{showcolor}{false}  

\newcommand{\changed}[1]{%
  \ifbool{showcolor}{\textcolor{black}{#1}}{#1}%
}

\newcommand{\changedmore}[1]{%
  \ifbool{showcolor}{\textcolor{black}{#1}}{#1}%
}

\newcommand{\changedmost}[1]{%
  \ifbool{showcolor}{\textcolor{blue}{#1}}{#1}%
}

\usepackage{microtype}
\renewcommand{\baselinestretch}{0.93}

\begin{document}

\title{
Field Calibration of Hyperspectral Cameras for Terrain Inference
}

\author{Nathaniel Hanson$^{1,2*}$, Benjamin Pyatski$^{1}$, Samuel Hibbard$^{1}$, Gary Lvov$^{1}$, Oscar De La Garza$^{1}$, \\ Charles DiMarzio$^{1}$, Kristen L. Dorsey$^{1}$, and Taşkın Padır$^{1,3}$
\thanks{Manuscript received: May 23, 2025; Revised August 12, 2025; Accepted September 16, 2025.}
\thanks{This paper was recommended for publication by Editor Lucia Pallottino upon evaluation of the Associate Editor and Reviewers' comments.}
\thanks{This research was sponsored by the United States Army Corps of Engineers (USACE) Engineer Research and Development Center (ERDC) Geospatial Research Laboratory (GRL) and was accomplished under Cooperative Agreement Federal Award Identification Number (FAIN) W9132V-22-2-0001. The views and conclusions contained in this document are those of the authors and should not be interpreted as representing the official policies, either expressed or implied, of USACE EDRC GRL or the U.S. Government. The U.S. Government is authorized to reproduce and distribute reprints for Government purposes notwithstanding any copyright notation herein.
}
\thanks{$^{1}$Nathaniel Hanson, Benjamin Pyatski, Samuel Hibbard, Gary Lvov, Oscar De La Garza, Charles DiMarzio, Kristen L. Dorsey, and Taşkın Padır are with the Electrical and Computer Engineering Department, Northeastern University, Boston, Massachusetts, USA.}
\thanks{$^{2}$Nathaniel Hanson is with the Lincoln Laboratory, Massachusetts Institute of Technology, Lexington, Massachusetts, USA.}
\thanks{$^{3}$Ta\c{s}k{\i}n Pad{\i}r holds concurrent appointments as a Professor of Electrical and Computer Engineering at Northeastern University and as an Amazon Scholar. This paper describes work performed at Northeastern University and is not associated with Amazon.}
\thanks{$^{*}$Correspondence: {\tt\footnotesize nathaniel.hanson@ll.mit.edu}}}
\markboth{IEEE Robotics and Automation Letters. Preprint Version. Accepted September, 2025}
{Hanson \MakeLowercase{\textit{et al.}}: Field Calibration of Hyperspectral Cameras for Terrain Inference}

\maketitle

\begin{abstract}
Intra-class terrain differences such as water content directly influence a vehicle’s ability to traverse terrain, yet RGB vision systems may fail to distinguish these properties.
Evaluating a terrain’s spectral content beyond red-green-blue \changedmost{wavelengths} to the near infrared spectrum provides useful information for intra-class identification. However, accurate analysis of this spectral information is highly dependent on ambient illumination.
We demonstrate a system architecture to collect and register multi-wavelength, hyperspectral images from a mobile robot and describe an approach to reflectance calibrate cameras under varying illumination conditions.
To showcase the practical applications of our system, HYPER DRIVE, we demonstrate the ability to calculate vegetative health indices and soil moisture content from a mobile robot \changedmost{platform}.
\end{abstract}



\section{Introduction}
\label{sec:intro}
\IEEEPARstart{U}{nderstanding} \textit{where} to drive is a critical question for autonomous vehicles~\cite{islam2022off}. Many systems operate by identifying and defining safe regions for traversability such as road, grass, soil, and sand from RGB camera images~\cite{schmid2022self}.
While broad terrain labels are useful in semantic segmentation, they do not capture intra-class differences---an oil slick on a road, ice on soil, or water-logged grass---that may impact vehicle traversability.
\changedmost{Adding spectral data as a complement to RGB images improves class separation} \cite{hanson2022hyperbot}, \changedmost{so incorporating} intra-class variations into robot intelligence \changedmost{may similarly} improve \changedmost{performance}. To realize this vision---dynamically identifying these differences onboard a field robot---new sensing and algorithmic approaches are required.

Hyperspectral imaging (HSI) is well-established in remote sensing for airborne terrain monitoring, material detection \cite{manolakis2013detection}, and anomaly identification \cite{matteoli2010tutorial}.
\changedmost{In contrast to the three intensity values present in RGB cameras,  hyperspectral imagers record dozens of discrete wavelengths.}
\changedmost{Characteristic absorbance peaks across the visible-to-infrared spectrum due to water, organic compounds, and other chemical compositions will then provide classification insight.}
While HSI often improves inter-class separability over RGB \changedmost{alone} \cite{hanson2022hyperbot}, its use has been constrained to environments with well-controlled illumination or frequent camera re-calibration using a target. These constraints are untenable for robots in the field.

\begin{figure}[!tbp]
    \centering
    \includegraphics[width=\linewidth]{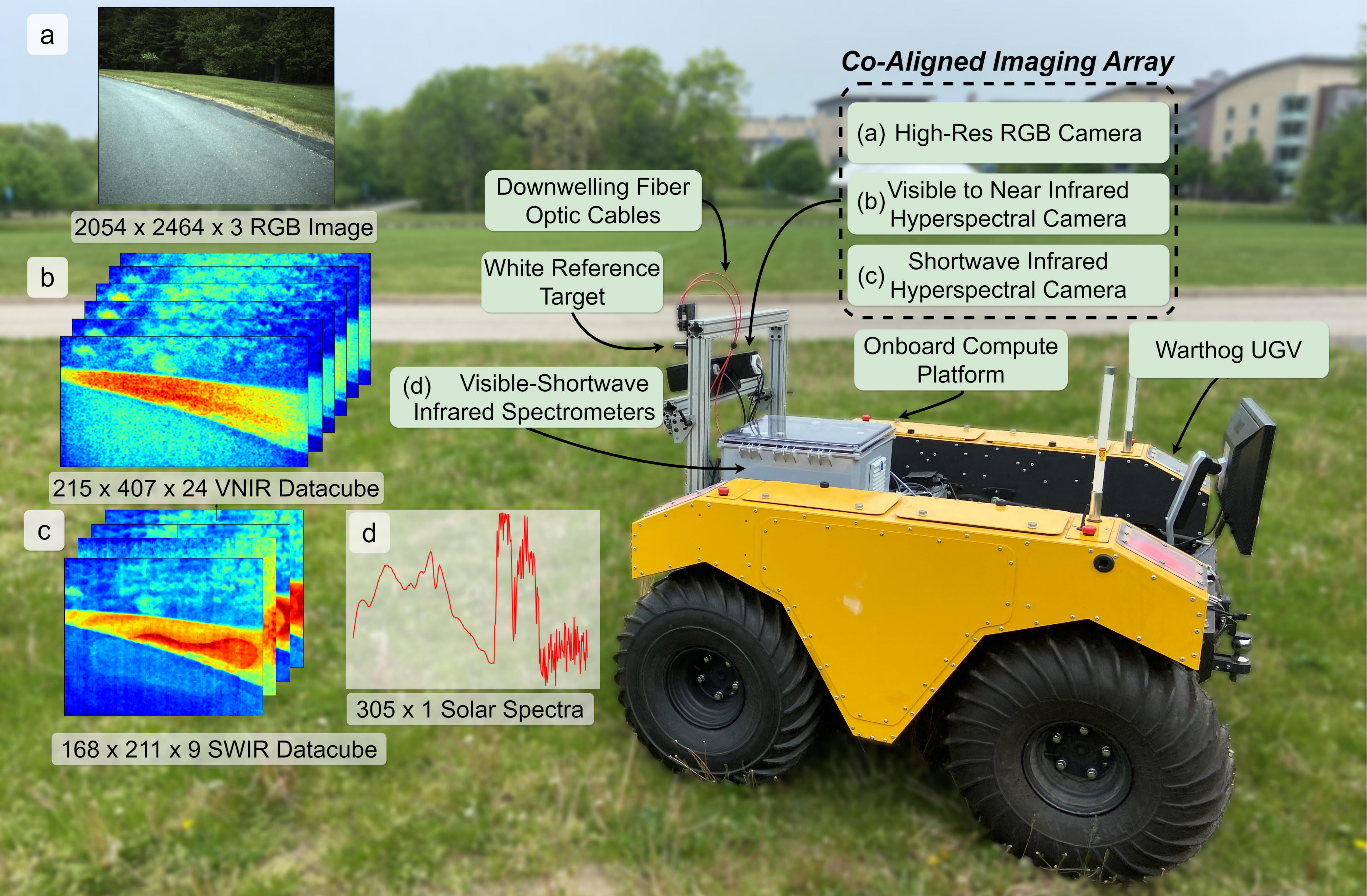}
    \caption{Our hyperspectral imaging system, HYPER DRIVE, mounted to an off-road mobile robot. Images are collected from (a) a high resolution RGB camera, (b) a visible-near infrared (VNIR) hyperspectral camera, and (c) a shortwave infrared (SWIR) hyperspectral camera. We use combined (d) point spectrometers to measure a white-reference signal and calibrate the hyperspectral cameras.}
    \vspace{-0.5em}
    \label{fig:hyper_drive_teaser}
    \vspace{-1.0em}
\end{figure}

\begin{figure*}[!t]
    \centering
    \vspace{0.5em}
    \includegraphics[width=\linewidth]{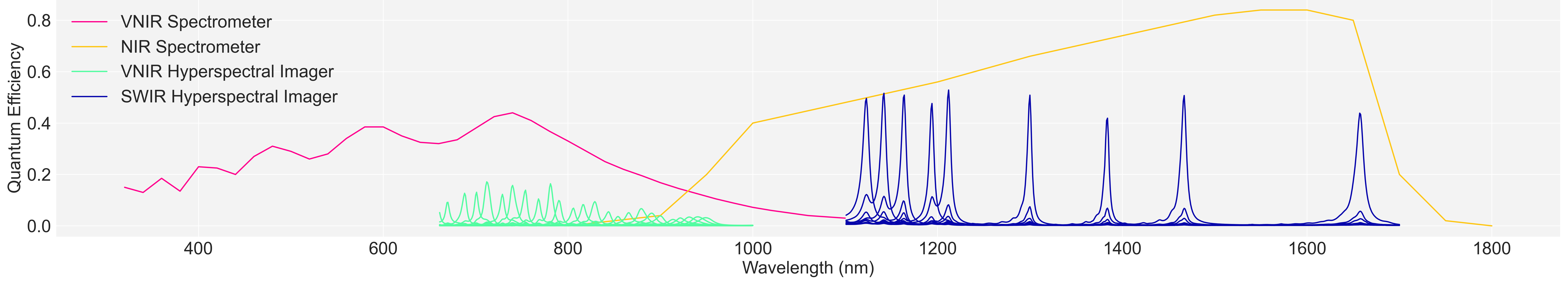} 
    \caption{\changedmost{The} likelihood of incident photon eliciting a digital count and sampling widths of \changedmost{the spectral sensors in HYPER DRIVE.}}
    \label{fig:spectral_sensitivities}
    \vspace{-1.5em}
\end{figure*}

We propose to use field robot-based HSI to identify intra-class terrain variations and dynamically calibrate changes in illumination conditions that would otherwise distort the spectral information. We apply HSI to identify intra-class terrain variations and demonstrate a system, HYPER DRIVE, to capture datacubes (i.e., a multi-dimensional view across spectrum and space) from a moving platform (Fig.~\ref{fig:hyper_drive_teaser}). We develop and validate a data-driven approach to dynamically calibrate the imaging areas via a learned up-sampling network that references a calibration target continuously sampled by an on-board point spectrometer.
Our work shows that forward-facing HSI enables low computational cost differentiation of visually similar terrain and is practical for field robotics.

This work contributes the following to the state-of-the-art:
\begin{itemize}
\item The design and development of HYPER DRIVE, a field robot specific hyperspectral camera system,
\item A method for spectrometer-hyperspectral joint calibration under varying illumination conditions, and
\item The evaluation of HYPERDRIVE’s performance in estimating soil moisture content and vegetative health.
\end{itemize}

\section{Background}
\label{sec:related_work}

\subsection{Hyperspectral imaging}
Hyperspectral cameras measure light intensity at multiple wavelengths in the visible ($400-800$ nm), the visible to near infrared (VNIR, $400-1100$ nm), and the shortwave infrared (SWIR, $1100-2500$ nm) \changedmost{spectra}.
The material composition of a scene may be identified from HSI through the spectral absorbances of various materials at specific wavelengths (i.e., their spectral ``signatures'').
For intra-class terrain identification, wavelengths related to water absorption ($970$ nm \cite{nann1991solar}, $1200$ nm, and $1400$ nm \cite{yamanouchi1985absorption}), and chlorophyll concentration (730 nm, \changedmost{e.g., from plants}), are critical.

\subsection{Related work}
Under controlled settings, hyperspectral sensors are calibrated to \changedmost{measure} spectral radiance. This approach is less practical in dynamic and autonomous systems because measuring outdoor radiance requires knowing atmospheric conditions and the position of the sun  \cite{ray1994faq}.
To address this challenge, recent work has scaled spectral readings between spectroradiometers mounted on ground-based and unmanned aerial vehicles (UAVs) \cite{hakala2013spectral} and reconstructed reflectance through a subspace representation and periodic imaging of a target \cite{wendel2017illumination}. This approach requires a robot to always be within navigable distance of a calibration target or to carry a large calibration target onboard.
\cite{khanna2017field} demonstrated on-robot calibration for a VNIR snapshot camera, but significant changes in the incident illumination required additional collection and calibration.
This approach also assumes that spectral distributions are constant over the VNIR range \cite{blackburn2012spectral}, \changedmost{which} does not hold in the SWIR range.

In autonomous systems, a more practical approach is to calculate the spectral reflectance, $R$, as a proportion of detected light relative to maximum signal intensity.
Following this approach, an HSI-specific version of min-max normalization adapted from \cite{geladi2004hyperspectral} is
\begin{equation}
    \label{eqn:standard_reflectance}
    \text{reflectance} = \frac{\text{signal} - \text{signal}_{\text{dark}}}{\text{signal}_{\text{reference}} - \text{signal}_{\text{dark}}}
\end{equation}
where the dark signal is the reading when the camera is covered, and the reference signal is the camera reading from a target. Typically, compressed polytetrafluoroethylene (PTFE) is chosen as the reference target due to its uniform reflection of visible - SWIR radiation.

Several previous works have demonstrated \changedmost{HSI} in robotics, albeit with assumptions about the respective camera's orientation or working distance. \cite{hostrand2019uav} demonstrated hyperspectral imaging on a UAV through an initial spectral calibration. This approach is less feasible for ground vehicles due to the large relative motion of the sun and the presence of other light sources, reflections, and objects on the ground.
\cite{hanson2022vast} demonstrated terrain classification by measuring the ground spectral signature near the tire contact point on a wheeled vehicle, but this approach required (i) a short distance (5 cm) between the spectrometer and the ground and (ii) active illumination. Integrating HSI across a range of autonomous vehicles will require large field of vision cameras rather than point spectrometers, dynamic and computationally feasible calibration strategies, and the use of only passive illumination. To the best of our knowledge, no work has addressed the challenge of dynamic calibration of hyperspectral cameras and generalizes \changedmost{across} camera models and wavelengths.

\section{System Architecture}
\label{sec:system_arch}
\subsection{Imaging System}
This work employs the \changedmost{HSI} system first developed in \cite{hanson2023hyperdrive}, which featured broader wavelength sensitivity than any current robot system and explored the use of non-visible light to understand broad inter-class terrain classes.
The sensor array includes hyperspectral cameras and a high-resolution RGB camera paired with a down-welling spectrometer to measure reflected solar light. The VNIR camera (XIMEA SNAPSHOT NIR, IMEC) captures $660-900$ nm using Fabry–Pérot filters across 24 spectral bands, while the SWIR camera (SNAPSHOT SWIR 9 Band, IMEC) captures $1100-1700$ nm with 9 spectral bands. Combined, the system provides 33 channels over 1100 nm of the spectrum. The cameras are housed in a weatherized 3D-printed (Onyx, Markforged) casing with broadband transmissive windows (Gorilla Glass, Corning) for minimal light perturbation. The VNIR and SWIR hyperspectral cameras have fields of view (FOV) of $25^{\circ}$. The complete sensor package, including housing, measures 250 $\times$ 82 $\times$ 73 mm, and has a mass of $\approx$ 5.0 kg. Data acquisition is handled by an onboard computer (Intel NUC 11, Core i7 Processor, 64 GB RAM, 2 TB SSD, NVIDIA RTX 2060 GPU), making the system practical for integration onto other UGVs.

The hyperspectral cameras are spatially registered with a 5 MP RGB machine vision camera with an FOV of $30.9^{\circ}$. A combined visible-shortwave infrared (VIS-SWIR) datacube is created using projective transforms and keypoint mapping. The final datacube ($1012 \times 1666 \times 33$) is cropped to overlapping regions and interpolated to match the RGB resolution.

\subsection{Point Spectrometers}
Complementing the cameras are two spectrometers covering $500-1700$ nm. The VNIR spectrometer (Pebble VIS-NIR, Ibsen Photonics) captures $500-1100$ nm with a silicon detector and 256 spectral pixels, and the SWIR spectrometer  (Pebble NIR, Ibsen Photonics) spans $950-1700$ nm with an InGaAs detector and 128 spectral pixels. Their overlapping range compensates for decreased efficiency at higher wavelengths and ensures enhanced spectral sensitivity. Fiber optic cables (Thor Labs) couple light into the spectrometers from ferrules offset 4 cm above a 99\% Spectralon white reference target (Labsphere). The quantum efficiencies of the devices are plotted in Fig.~\ref{fig:spectral_sensitivities}. These spectrometer readings are used to dynamically generate reflectance calibrations according to the method discussed in Section~\ref{sec:white_ref_algorithm}.

\section{Spectral White Reference Generation}
\label{sec:white_ref_algorithm}
\subsection{Data Collection}

\begin{figure}[!t]
    \centering
    \vspace{0.5em}
    \includegraphics[width=\linewidth]{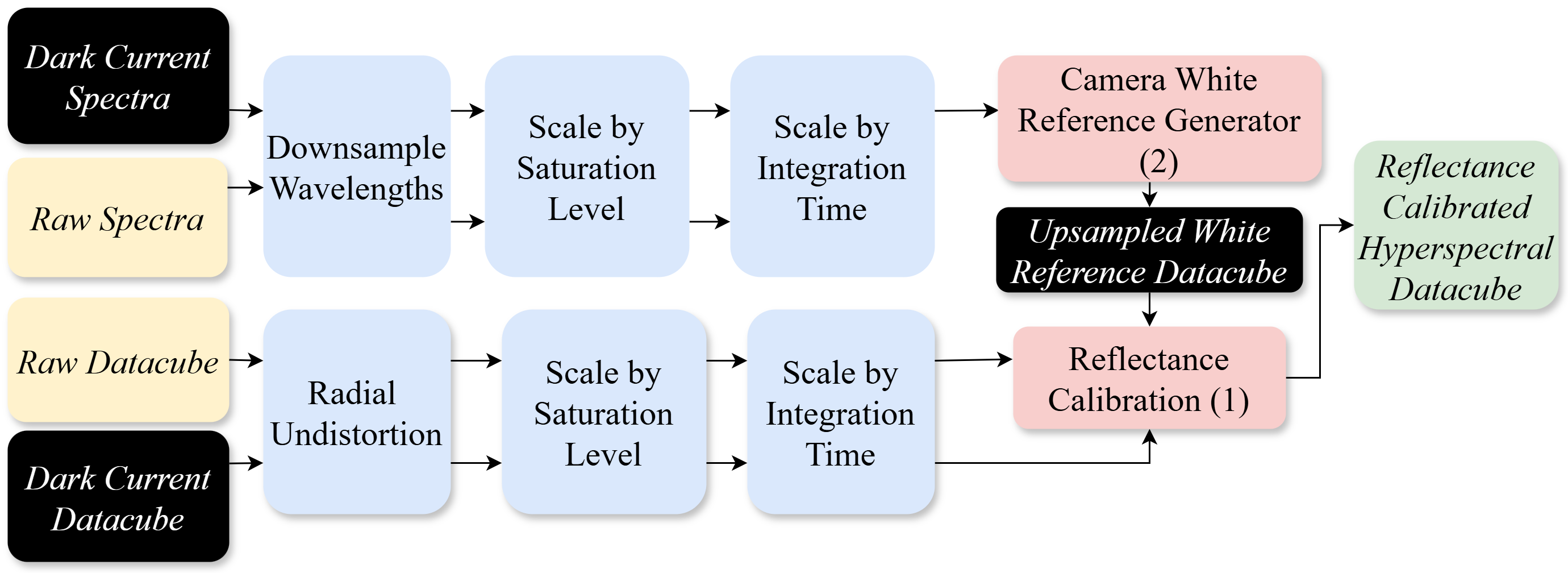} 
    \caption{\changedmost{To calibrate a datacube, the workflow takes in raw HSI hypercube data and spectral data from the down welling spectrometer  (yellow), compares it to static calibration references (black), transforms it (blue), and passes it to the relevant algorithms (red).}}
    \label{fig:calibration_flowchart}
    \vspace{-1.5em}
\end{figure}

To relate point spectrometer readings to hyperspectral images, we designed a set of experiments to measure the correspondence between the devices when they are viewing a calibration target made of the same material. The system was placed outdoors and aligned to the $5\degree$ azimuth with the solar track. This placement also ensured that no external shadows fell on either the spectrometer target or on the white reference target.
As the solar track varies only slightly from one day to the next, we assumed that the maximum solar overhead angle was constant across the measurements.

Across three days, the system \changedmost{was }operated with the parameters in Table~\ref{tab:training_parameters} under low to full sunlight from early morning to early evening and varying precipitation conditions. Devices sensing in the SWIR domain typically have longer integration times than VNIR devices due to decreased photon energy and proportionally lower atmospheric transmission.

\begin{table}[!b]
    \caption{HYPER DRIVE Training Device Parameters}
    \label{tab:training_parameters}
    \centering
    \footnotesize
    \setlength\tabcolsep{4 pt} 
    \begin{tabular}{c|c}
    \hline
    \textbf{Parameter} & \textbf{Value} \\
    \hline
    VNIR HSI Frame rate        & 10 Hz \\
    VNIR HSI Integration Time & 0.5 ms \\
    SWIR HSI Frame rate         & 10 Hz \\
    SWIR HSI Integration Time  & 1.0 ms \\
    VNIR Spectrometer Integration Time & 0.5 ms \\ 
    SWIR Spectrometer Integration Time & 50 ms \\  
    \hline
    \end{tabular}
\end{table}

A synchronized sample was acquired from each device every 5 minutes and saved with the time stamp at which it was acquired. In total, more than 2500 samples were collected to form a set of calibration spectra $P$.

\subsection{White Reference Generation Formulation}
The terms in Table~\ref{tab:gen_parameters} define hyperspectral and spectrometer data terms that are relevant to our analysis. For all variables, the subscript indicates the type of device (spectrometer $S$ or HSI $I$) and the wavelength range (VNIR or SWIR). In the general case, we desire to approximate a function $f$ (\ref{eq:stoi}), which maps an $S$ to $I$ as:
\begin{equation}
I = f(S)
\label{eq:stoi}
\end{equation}

\begin{table}[!tb]
    \vspace{0.5em}
    \caption{Targetless White Reference Generation Parameters}
    \label{tab:gen_parameters}
    \footnotesize
    \setlength\tabcolsep{4 pt} 
        \begin{tabularx}{\linewidth}{M{0.25\linewidth}|M{0.65\linewidth}}
    \hline
    \textbf{Parameter}     & \thead{\textbf{Description}} \\ \hline
     $t$        & Integration time, the time for which the sensor was exposed to light.       \\ \hline
    $I\{H,W,\Lambda_{I}\}$ & 3D hyperspectral datacube with height $H$, width $W$, and wavelength channels $\Lambda$ in digital counts. Pixels of the datacube are indexed using the column $c$, row $r$, and channel $\lambda$. $I \in \mathbb{Z}^{H \times W \times \Lambda}+$         \\ \hline
    $S\{\Lambda_S\}$         & 1-D Vector, representing the digital counts of the spectrometer. $S \in \mathbb{Z}^{\Lambda_{S}}+$        \\ \hline
$D$  &  Bit depth of a device, where the subscript represents the specific device and range      \\ \hline
    $P$      & Set of all observed spectra         \\ \hline
$\Lambda$      & Set of wavelengths, subscript indicates the specific device or subset         \\ \hline
    \end{tabularx}
    \vspace{-2.50em}
\end{table}

As we have sets of devices that cover the wavelength spectrum, we consider the calibration of VNIR and SWIR devices separately. In this calibration routine, we assume integration times are set low enough to avoid saturation. At every time step, we assume there is a spectrometer reading of digital counts $S_{raw}$ and a hyperspectral datacube also of digital counts $I_{raw}$. The saturation point of the digital count for these devices varies widely by device; therefore, we first begin by scaling each measurement to be normalized in the range $[0,1]$ by dividing by the maximum digital count value at saturation. We denote by $S \in \mathbb{Z}^+$ and $I \in \mathbb{Z}^+$ the raw, unnormalized spectrometer and hyperspectral measurements in digital counts. The normalized quantities, $S_{\text{norm}}$ and $I_{\text{norm}}$, are obtained by scaling the raw counts to the range $[0,1] \subset \mathbb{R}^+$ according to (\ref{eq:norm}).

\begin{align}
S_{norm} = \frac{S_{raw}}{D_{S}}, \quad I_{norm} = \frac{I_{raw}}{D_{I}}
\label{eq:norm}
\end{align}

$D_S$, $D_I$ denote the device-specific saturation counts. These normalized real-valued quantities are the inputs to the mapping function $f$.

Due to differences in grating technologies and filter characteristics, the discrete center wavelengths measured by the hyperspectral camera and the spectrometer are not perfectly aligned. We map between the wavelengths of the two devices with the following operations. Let
\begin{align*}
\Lambda_S = {\Lambda_S^{(a)}}_{a \in A} \text{where each } \Lambda_S^{(a)} \in \mathbb{R}\\
\Lambda_I = {\Lambda_I^{(b)}}_{b \in B} \text{where each } \Lambda_I^{(b)} \in \mathbb{R}
\end{align*}
denote the sets of center wavelengths for the spectrometer and the camera, respectively, where \(A = \{1, \dots, N_S\} \) and \( B = \{1, \dots, N_I\} \) index the corresponding channels and $N_S$ and $N_I$ are the number of discrete channels for the spectrometer and hyperspectral camera, respectively.
For each \( b \in B \), we find the index of the spectrometer wavelength closest to the camera wavelength \( \Lambda_I(b) \)  and use it to define a mapping:

\begin{align}
a* = \argmin_{a \in A} | \Lambda_S^{(a)} - \Lambda_I^{(b)}| \\
\label{eq:closest_wave_3}
\Lambda_{cal}^{(b)} = \Lambda_S^{(a*)}
\end{align}

which aligns each camera wavelength with its nearest neighbor in the spectrometer domain. (\ref{eq:closest_wave_3}) downsamples the spectrometer signal to the same number of wavelength values as the corresponding hyperspectral camera. This operation presupposes that the range of $\Lambda_S$ spans the minimum and maximum values of $\Lambda_I$ in order to have good correspondences. In all further mentions of a spectral signal $S_{raw}$, we assume this signal is already subsampled to the dimension of $\Lambda_{cal}$, a subset of wavelengths corresponding between the two devices. Fig.~\ref{fig:pixel_based_modeling}a depicts this downsampling operation.

Next, we subtract off the device dark counts. These counts are acquired when the device is capped and no photons are able to enter the sensing aperture. Thus, the observed reading is an estimate of the intrinsic sensor noise. In practice, this value is found to be stable once the device reaches a steady-state operating temperature. Taking the minimum values at each wavelength yields an appropriate lower bound. Similarly, the dark counts are also normalized with respect to the digital counts. These values are then clamped to $\geq 0$.
\begin{align}
\label{eq:Snormz}
S_{norm,z} = \text{max}\left(0, S_{norm}-\frac{\text{min}(S_{dark})}{D_{S}}\right)\\
I_{norm,z} = \text{max}\left(0, I_{norm}-\frac{\text{min}(I_{dark})}{D_{I}}\right)
\label{eq:Inorm_z}
\end{align}
To generate our approximation of $f: S\rightarrow I$, we expand the domain of this function to include the domain of the normalized spectra $S_{norm,z}$, and the location of the pixel on the photodetector array, $r,c$. As shown in \cite{khanna2017field}, snapshot hyperspectral cameras tend to have non-negligible optical attenuation, especially near the edges of the image. For now, we will leave this function denoted by the generic $f(S_{norm,z},r,c)$.

To construct the white reference for the camera, we utilize the following formulation:

\begin{align}
    \label{eq:I_basic}
 I_{white}(r,c) = f(S_{norm,z},r,c)\\
\nonumber\text{where } r \in \{0,...,H\} \text{ and } c \in \{0,...,W\}
\end{align}

(\ref{eq:I_basic}) yields a white reference cube in the shape of the target hyperspectral datacube derived from the spectrometer white reference spectra. We are now able to estimate the calibrated hyperspectral datacube $I_{cal}$, using (\ref{eqn:standard_reflectance}), but generalized to hyperspectral datatypes.
\begin{equation}
    I_{cal} = \frac{\frac{I_{raw}}{D_I} - \frac{I_{dark}}{D_I}}{I_{white}-\frac{I_{dark}}{D_I}}
    \label{eq:ical}
\end{equation}
In (\ref{eq:ical}), the integration time from the calculation of the white reference is excluded. Considering that this data was collected with a fixed integration time $t_{base}$, we simply consider a scale factor given as the ratio of the base time to the current time. This is motivated under the assumption that the digital counts increase linearly with changes in the integration time.
\begin{equation}
    t_{scale} = \frac{t_{base}}{t_{new}}
    \label{eq:tscale}
\end{equation}
The scale factor from (\ref{eq:tscale}) is needed whenever the integration time of the spectrometer or hyperspectral camera differs from the original set integration time. Adding an integration scale factor yields (\ref{eq:spec_hsi_joint_calibration}), which will be used to recover the reflectance datacube. This equation is still dependent on the mapping function $f$, which we will learn an approximate solution to using empirical data in the following section. Graphically, this calibration process is shown in Fig.~\ref{fig:calibration_flowchart}.

\begin{equation}
    \label{eq:spec_hsi_joint_calibration}
    \resizebox{\linewidth}{!}{%
    $I_{cal} = \frac{(\frac{I_{raw}}{D_I}  \times t_{scale,I}) - (\frac{I_{dark}}{D_I}\times t_{scale,I})}
    {f((\frac{S_{raw}}{D_{S}} - \frac{S_{dark}}{D_{S}})\times t_{scale,S},\forall r,\forall c)-(\frac{I_{dark}}{D_I}\times t_{scale,I})}$
    }
\end{equation}

\subsection{Modeling Spectral Mapping Functions}
\begin{figure}[!tbp]
    \centering
    \vspace{0.5em}
    \includegraphics[width=\linewidth]{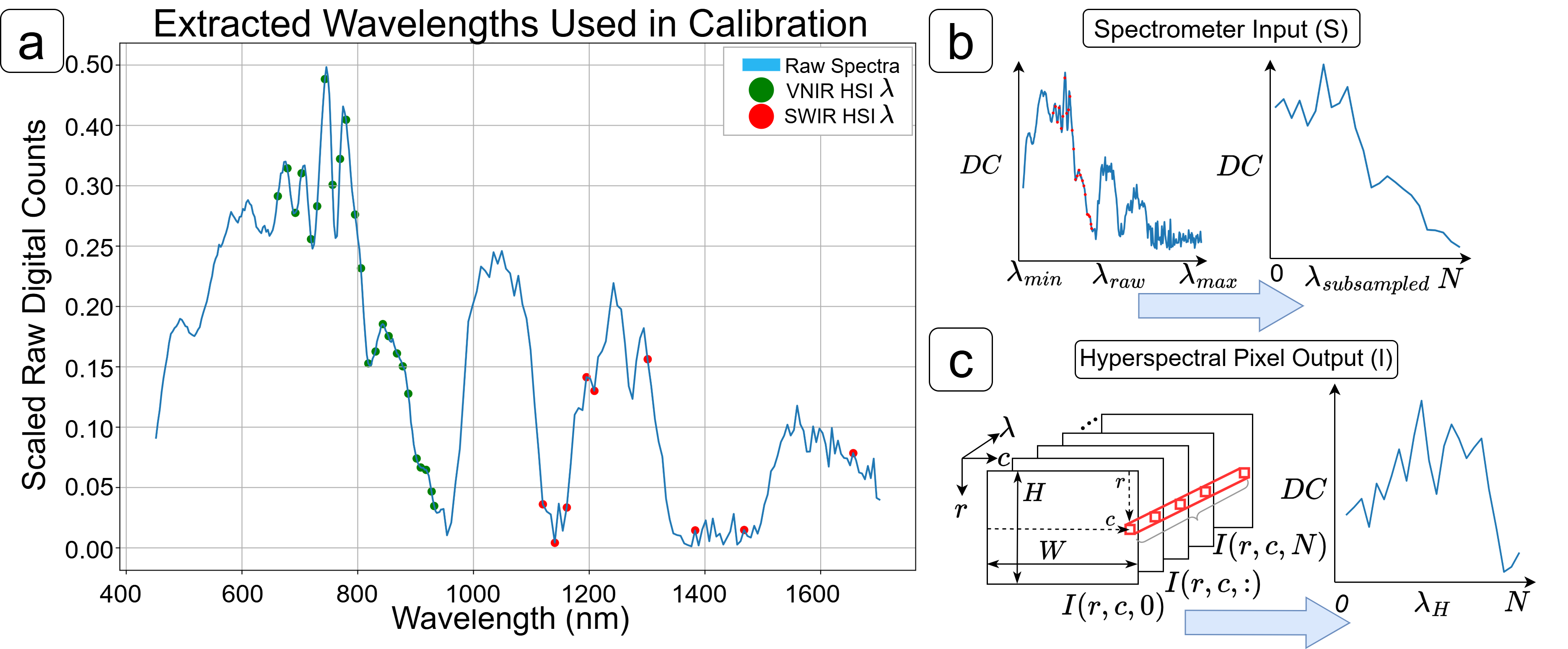} 
    \caption{(a) Single digital number normalized white reference measure from the HYPER DRIVE spectrometer system with the sampled spectra of the VNIR HSI (green dots) and SWIR HSI (red dots). (b) Downsampling of raw spectra to the wavelengths associated with the (c) temporally synchronized hyperspectral image.}
    \label{fig:pixel_based_modeling}
    \vspace{-1.50em}
\end{figure}

Given the observations of the correlation, or lack thereof, between spectra, we attempt to fit two different kinds of regression models on a per-pixel basis, using a) Multiple Linear Regression (MLR) and b) Multi-Layer Perceptron (MLP) Regression. We selected these two approaches since they constitute linear and non-linear modeling to the white reference reconstruction problem. We first begin by introducing the generic formulation of the problem. These formulations are applied to each wavelength pairing of the spectrometer and hyperspectral camera. Given its ability to handle spatial context, a more complex convolutional neural network (CNN) outwardly appears to be well-suited to this reconstruction task. Our experimentation with these CNN-based models showed a lack of convergence, likely due to requirements for a larger dataset to train a model of this complexity. Hence, we adhered to a per-pixel reconstruction of white reference data.
\paragraph{Model Inputs \& Outputs}
We define the input vector to this problem for a single sample as the reduced order spectral signal as calculated with (\ref{eq:closest_wave_3}). From a complexity perspective, downsampling also greatly helps save on computation time and model storage. Note that in the following equation, we only consider the output of a single pixel at a time. Our modeling efforts seek to find a model which associates a set of input spectra with $\Lambda_I$ wavelength values, with the pixel response at location $(r,c)$ that also contains $\Lambda_I$ values. For the general case, we consider there are $P$ available samples from the rooftop collection data. The problem is modeled graphically in Fig.~\ref{fig:pixel_based_modeling}. We also assume that all values of $S_p$ and $I_p$ are $\in \mathbb{R}+$.

\begin{table*}[bp]
\vspace{-0.5em}
\caption{Summarized Reconstruction Error Statistics}
\label{tab:reconstruction_stats}
\centering
\footnotesize
\setlength\tabcolsep{4 pt} 
\begin{tabular}{c|c|c|c|c|c|c|c|c|c}
\toprule
Model Type & Sensor & $\overline{MSE}$ & $\sigma(MSE)$ & $\overline{MAE}$ & $\sigma(MAE)$ & $\overline{SAM}$  & $\sigma(SAM)$ & Model Size/Pixel (MB) & Inference (s)\\
\midrule
\multirow{2}{*}{MLR} & VNIR & 0.0001 & 0.0000 & 0.0041 & 0.0006 & 0.2380 & 0.0272  &  0.0016 & 3.41\\
 & SWIR & 0.00000 & 0.0000 & 0.0008 & 0.0001 & 0.3376 & 0.0248 & 0.0018 & 1.38\\
\multirow{2}{*}{MLP} & VNIR & 0.0019 & 0.0007 & 0.0290 & 0.0062 & \bf{0.1807} & 0.0496 & 0.0299 & 5.34\\
 & SWIR & 0.0029 & 0.0018 & 0.0324 & 0.0137 & \bf{0.2477} & 0.0556  & 0.0294 & 2.26\\
    \bottomrule
    \end{tabular}
\end{table*}


\paragraph{Regression Modeling}
\label{sec:hyperdrive:regression}

The first way of conceptualizing the relationship between input and output spectra is that of a multi-output regression model. Given the dataset collected during the rooftop system calibration campaign, we have a set of spectrometer measurements, each correlated with one of the white reference measurements taken for a given cube. MLR attempts to learn a weight matrix with ordinary least squares to map between the two sensors with minimal error.



\paragraph{Multi-Layer Perceptron}
We also considered a deep learning model based on an MLP to approximate $f$. The general structure of these models followed an encoder-decoder architecture, where the encoder learns to encode an input signal to a reduced-dimension latent space before reprojection to a higher dimension. The learned network consisted of 4 fully-connected layers with 24, 10, 10, and 24 neurons, respectively. A ReLU function is placed after each of the fully-connected layers.

When training these models, the loss term is crucial in accurately reconstructing the white reference hypercube. For instance, a signal that is reconstructed with similar absorbance troughs and peaks, but at lower magnitude is more desirable than one with a low mean-squared error (MSE) but misplaces the location of spectral features. The use of spectral angle mapper (SAM) is well supported by the literature as a means to calculate the divergence between two spectral signals \cite{kruse1993spectral}. We construct a composite loss function which simultaneously considers SAM loss and MSE loss. The equations for MSE (\ref{eq:mse}) and SAM (\ref{eq:sam}) specific to this task are presented below.    
\begin{equation}
\mathcal{L}_{MSE}(r,c) = \frac{1}{P} \sum_{p=1}^P(I_{p}(r,c) - I_{p}^{'}(r,c))^2
    \label{eq:mse}
\end{equation}
\vspace{-2.00em}
\begin{multline}
    \label{eq:sam}
\mathcal{L}_{SAM}(r,c) = \\ \arccos \left(
\frac{ 
\sum_{p=0}^P I_{p}(r,c) \cdot I_{p}^{'}(r,c)
}{
\sqrt{ \left( \sum_{p=0}^P I_{p}(r,c) \right)^2 }
\sqrt{ \left( \sum_{p=0}^P I_{p}^{'}(r,c) \right)^2 }
}
\right)
\end{multline}
The range of $MSE \in \mathbb{R}+$ ;  $SAM \in [ 0,\pi] $. Together, the loss functions are combined with a weighting term $ \alpha $ to weight their contributions to the total loss for the current sample. For training these models, $\alpha$ is set to 0.1. The combined loss function is given in (\ref{eq:pixel_reconstruction_loss}).
\begin{equation}
    \label{eq:pixel_reconstruction_loss}
    \mathcal{L}_{total} = \mathcal{L}_{MSE} + \alpha \mathcal{L}_{SAM}
\end{equation}
As the models are fitted on a per-pixel basis, they are trained in parallel. GPU acceleration is not needed for the MLP training as loading data onto the GPU for each model training process often takes longer than the convergence time of the model on the CPU. Models are implemented with the Scikit-Learn library \cite{scikit-learn} and trained for 1000 epochs, with early stopping if the loss function does not improve by $0.0001$ units within 10 successive epochs. The Adam optimizer \cite{kingma2014adam} is used to update the weights during training.

\paragraph{Data Augmentation \& Processing}
The spectral data was augmented in two ways to increase the generality of the learned models. First, we replicated the data by a factor of 3. We then applied scaling noise to both the input and output data drawn from a uniform distribution, adding up to $\pm 10\%$ scaling: $s_p = (1+\mathcal{U}(-0.1,1.0)) * s_p$, where $\mathcal{U}$ is a uniform probability distribution. We clipped the data to ensure no negative values were introduced. Secondly, with an event probability of $\mathbb{P}=0.10$, random indices in the range $\{0,...,|\Lambda_{cal}|\}$ of both the input and output were zeroed. This strategy was used as a means to encourage the model to learn generalized weights that handle different spectral distributions. This data was then split into 80/10/10\% training, validation, and testing percentages, respectively. 

\section{Modeling Results}
We assess the performance of MLP and MLR models using generalized SAM and MSE metrics as error measures rather than loss functions. Metrics such as mean absolute error (MAE) and MSE are calculated per pixel and averaged over all samples in $P_{test}$, where $I$ is the target and $I^{'}$ represents model predictions. The VNIR MLR model exhibits the lowest errors near the center of the detector, with errors increasing radially outward, while the SWIR camera shows circularly radiating errors due to intrinsic system noise.

MLP generally overestimates white spectra intensity compared to MLR but captures spectral troughs and peaks more accurately, aiding appropriate scaling. Overestimation is acceptable if consistent, as underestimation can result in target pixel values exceeding the $[0,1]$ normalization range, leading to excessive clipping and loss of spectral detail. For scaled spectrum inputs below 10\% of the device’s dynamic range, MLR outperforms MLP in accuracy. However, MLP achieves reasonably correct magnitudes with some noise. Optimal model performance requires sufficient signal, suggesting future work on dynamically adjusting integration times.

Summary statistics in Table~\ref{tab:reconstruction_stats}, computed with the held-out test set, show that MLR yields lower MSE and MAE, aligning with its optimization via MSE. However, MLR exhibits greater spectral angle divergence (higher SAM values). In contrast, MLP models generate spectra with the same shape (characteristic peaks and troughs) as the reference spectra, with trade-offs in inference time and model size.

Given limited computational resources, models must balance storage and runtime efficiency. MLR models, with matrix-based weight storage, scale quadratically with input-output dimensions but are smaller and faster than MLPs. Though none of the models operate in real-time, stable ambient solar spectra reduce the need for ad-hoc white reference recalibration. This stability allows robotic systems to allocate hardware resources more effectively to other tasks, as the solar spectrum does not change on a per-minute basis.

\subsection{Color Marker Recovery}
\label{sec:hyperdrive:color_checker}

\begin{figure*}[!tb]
    \centering
    \vspace{0.5em}
    \includegraphics[width=\linewidth]{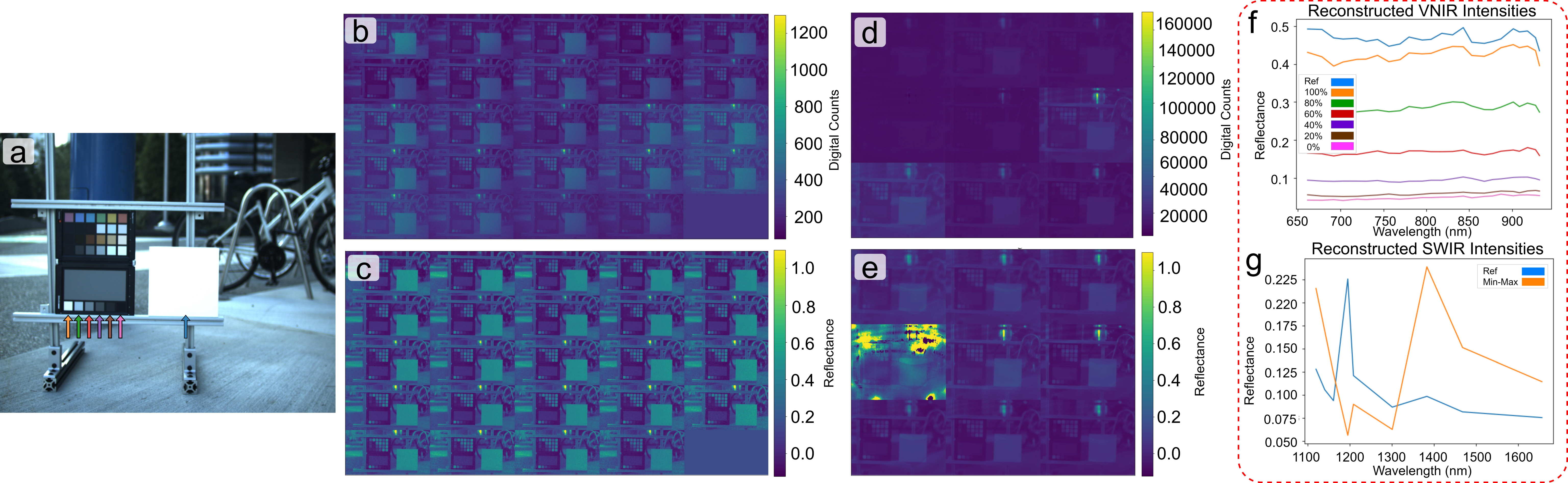} 
    \caption{(a) RGB image showing color checker (left) and Spectralon white reference target (right). The colored arrows correspond to the calibrated grayscale intensities seen in (f). (b) Raw demosaiced hyperspectral image with intensity in digital counts. Wavelengths increase from top-left to bottom right. (c) Reflectance calibrated image montage using the per-pixel MLP. Wavelengths increase from top-left to bottom right. (d) Raw digital count measurement from SWIR camera. (e) Reflectance calibrated image montage using the per-pixel MLP. Wavelengths increase from top-left to bottom right. (f) Reconstructed VNIR reflectance intensities for a 5-pixel neighborhood located in each target square. (g) Reconstructed SWIR reflectance intensities for Spectralon target in scene compared to naive min-max normalization.}
    \label{fig:color_checker}
    \vspace{-1.5em}
\end{figure*}

To evaluate the proposed method, we tested its ability to produce consistent measurements for reflectance targets of known intensity using a calibrated color checker (Checkr 24, Spyder). The setup, shown in Fig.~\ref{fig:color_checker}, includes the color checker and a Spectralon tile for establishing illumination ground truth. The bottom row of the checker contains grayscale tiles of decreasing intensity. Fig.~\ref{fig:color_checker}a highlights these tiles, and Fig.~\ref{fig:color_checker}c presents the calibrated image. Spectralon exhibits uniform intensity across the spectrum, and the grayscale tiles correctly decrease in intensity, albeit with the 20\% and 0\% intensities closer than expected due to the camera's poor low-light sensitivity. Calibration mitigates variations seen in the uncalibrated datacube (Fig.~\ref{fig:color_checker}b), stabilizing VNIR reference spectra to within 5\% of the mean reflectance.

Similar trends are observed in SWIR calibration. Uncalibrated SWIR data (Fig.~\ref{fig:color_checker}d) display significant intensity variation across wavelengths, while the calibration method (Fig.~\ref{fig:color_checker}g) normalizes these discrepancies. However, low ambient illumination at 1195 nm results in a mean signal of $4.5 \pm 2\%$, creating calibration challenges due to low intensity and noise. Despite this, the calibration normalizes the white reference to within $\pm 3\%$, except for a spike at 1195 nm due to the aforementioned absorption in this band. This aligns with expectations for uniform intensity recovery.

In both VNIR and SWIR reconstructions, spectral values are below the ideal value of 1 for the Spectralon tile. Variations in average spectral intensity arise from factors such as atmospheric water content or shadows, as noted in prior research \cite{tucker1980remote,carter1991primary}. Per-wavelength normalization remains consistent, as evidenced in Fig.~\ref{fig:color_checker}. Techniques such as SAM, commonly used in remote sensing and machine learning \cite{khanna2007development}, are robust against albedo variations, making them suitable for scenes with varying intensities.

\section{Terrain Parameter Estimation}
\label{sec:terrain_property}
With the properly calibrated images, we consider the estimation of two terrain properties: vegetative health (VH) and soil moisture content (SMC) from a moving autonomous vehicle.  For these experiments, the HYPER DRIVE system was mounted on a large field robot (Warthog, Clearpath Robotics) to collect outdoor imagery. When mounted, the camera array is 1.0 meter off the ground. Because the spectrometer samples a uniform white reference tile mounted rigidly to the platform, transient shocks or vibrations from UGV motion do not affect the calibration.

\subsection{Vegetative Health Analysis}
As a field robot is moving through an environment, understanding where vegetation is present can aid in gauging the feasibility of a selected path. For instance, if the target path is primarily composed of gravel or sand, vegetation can be considered a non-viable area to traverse as vegetation is not normally present in cleared paths.

Normalized difference vegetation index (NDVI) is a well-studied and proven metric towards understanding the location of vegetation \cite{rouse1974monitoring}. NDVI is a normalized comparison between two spectral reflectance bands with a range of $[-1,1]$. Dense, healthy vegetation will yield a value near 1, while unhealthy vegetation will be less than 0.5. Values less than 0 suggest the presence of standing water.

Using our camera system, the NDVI is calculated along the channel axis with (\ref{eq:ndvi}). The subscripts to $I$ represent the wavelength bands extracted for the calculation of the index.

\begin{equation}
    \label{eq:ndvi}
    NDVI = \frac{I_{901} - I_{661}}{I_{901} + I_{661}}
\end{equation}

With the NDVI values, we extract a statistical decision boundary between vegetation and non-vegetation using Otsu's method \cite{otsu1979threshold}. This method assumes there are exactly two classes and seeks to minimize intra-class variance for the binary decision problem. We prefer the binary Otsu over a multi-Otsu thresholding \cite{liao2001fast} implementation as we presume vegetation vs. non-vegetation is a useful differentiation for off-road navigation and has precedence in the literature \cite{khanna2017field}. No post-processing or contrast enhancements are applied to the NDVI scores.

\begin{figure}[!t]
    \centering
    \includegraphics[width=\linewidth]{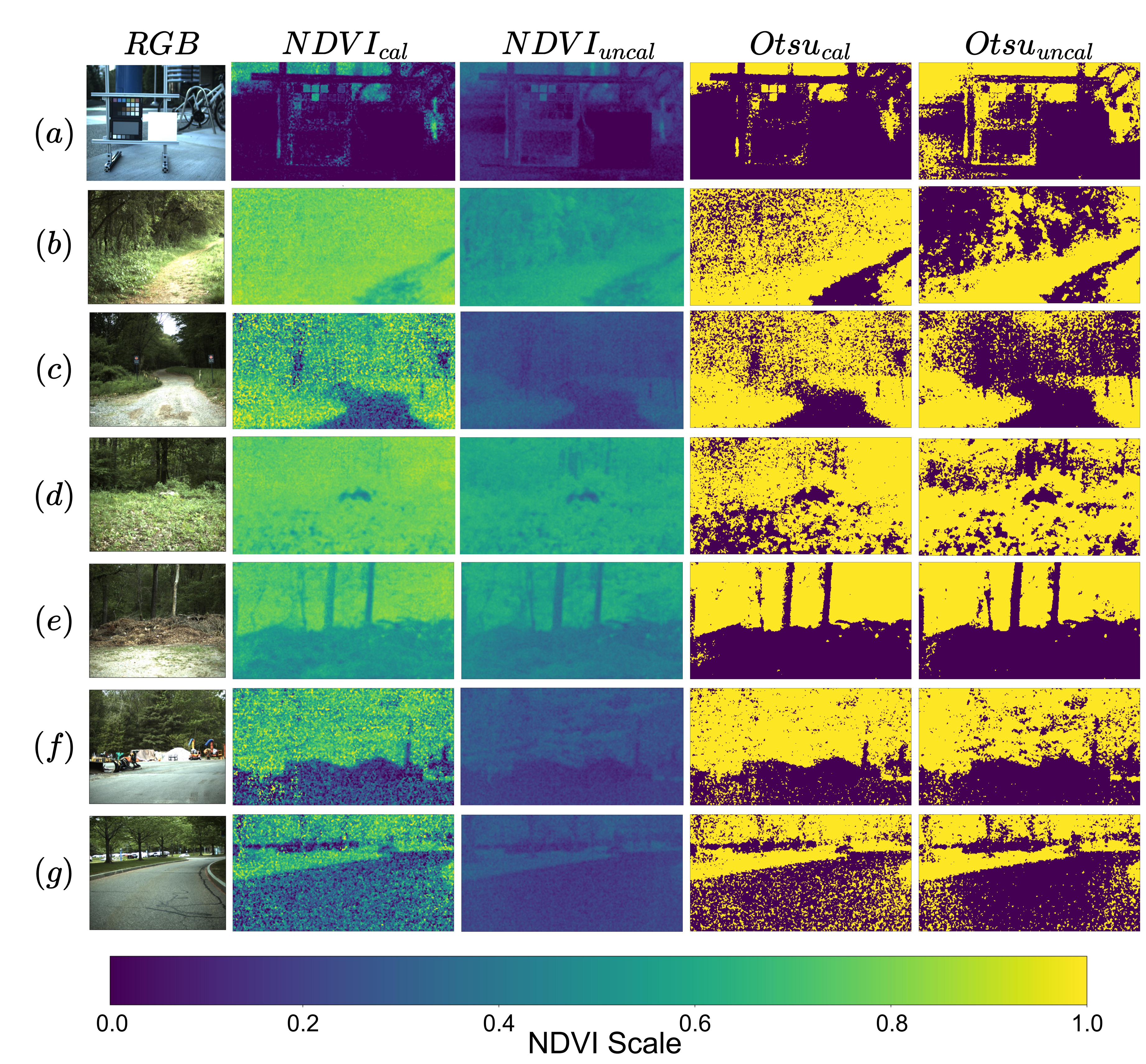} 
    \caption{NDVI values calculated using both the raw images and the proposed calibration method. These images are sampled from the HYPER DRIVE \cite{hanson2023hyperdrive} dataset.}
    \label{fig:ndvi_calculation}
    \vspace{-1.5em}
\end{figure}

The first column of Fig.~\ref{fig:ndvi_calculation} shows the raw RGB image for the scene. The next two columns show the calculated NDVI for both the raw and calibrated images. The final two columns show the Otsu segmentation performed on the calibrated and uncalibrated images, respectively. In the Otsu images, yellow indicates the pixels clustered as vegetation, while dark purple indicates non-vegetation.

\begin{figure*}[!t]
    \centering
    \vspace{0.5em}
    \includegraphics[width=\linewidth]{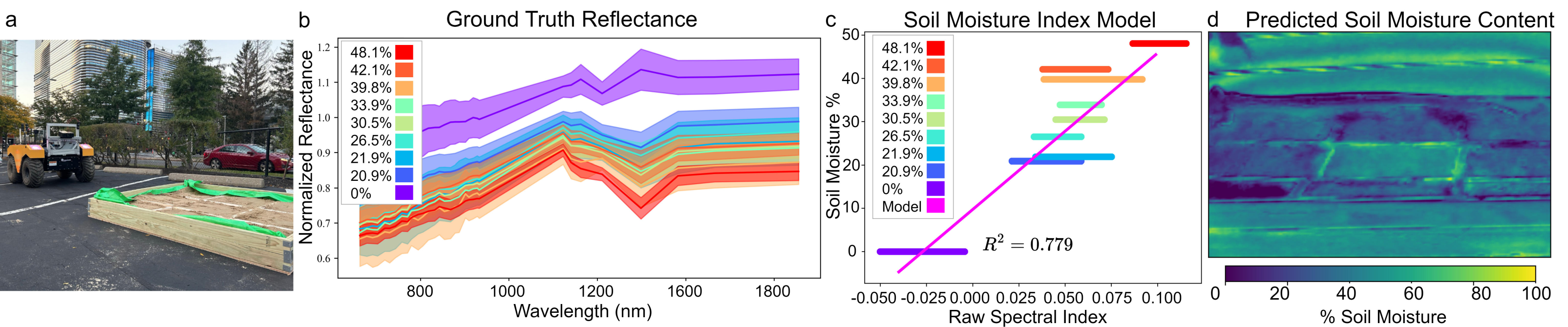} 
\caption{(a) HYPER DRIVE mounted to the field robot scanning sample testbed. (b) Ground truth spectral signatures for the calibration testbeds. (c) Soil moisture content in each testbed cell with fit regression line. (d) Predicted soil moisture using the regression model.}
    \label{fig:calibration_boxes}
    \vspace{-2.0em}
\end{figure*}
Row $(a)$ presents an RGB color checker scene with a background of vegetation and predominantly man-made structures. NDVI calculations reveal a strong contrast between vegetated areas and other regions. Reflective surfaces on the color checker edges produce non-zero NDVI readings. Otsu's method better segments vegetation compared to uncalibrated data, which has more pixels misclassified as vegetation. Rows $(b)$ and $(c)$, featuring dense undergrowth and shadows, exhibit lower raw HSI datacube intensity. Calibrated NDVI effectively handles illumination variations, though darker areas in uncalibrated data are misclassified. Some speckling in calibrated NDVI is noted due to low signal strength.

Rows $(d)$ and $(e)$ show varying vegetative health, with living plants and dry leaves. Calibrated Otsu improves detection of verdant vegetation and anomalies like boulders $(d)$. It also better identifies background trees as vegetation $(e)$, compared to uncalibrated data. Rows $(f)$ and $(g)$ feature mixed vegetation and man-made structures. Calibration stabilizes vegetation NDVI in low-light scenarios $(c),(f),$ and $(g)$, while uncalibrated NDVI values fluctuate with illumination.

Otsu's method provides insight into class separability but assumes only two distinct classes. In scenes with multiple classes, such as stressed vegetation or standing water, its heuristic nature may skew results. Nonetheless, our results show calibrated NDVI is a useful prior or input for semantic segmentation algorithms, enabling effective terrain analysis.

\subsection{Soil Moisture Detection}
Terrain moisture is also a significant area of interest to enable decision-making grounded in physical quantities. Towards understanding soil moisture content (SMC), we constructed a series of testbeds where soil moisture could be precisely controlled and measured in situ with probes. Fig.~\ref{fig:calibration_boxes}a shows the test setup in a gridded format of increasing soil moisture with HYPER DRIVE imaging the samples.

The testbed \changedmost{was} constructed in a $3 \times 3$ grid, with separations between adjacent soil samples to control the water content and minimize flow between cells. One box was kept dry while the remaining boxes had increasing amounts of water. Ground truth \changedmost{for SMC was} acquired by measuring soil relative humidity (RH) (\%) using a surface-penetrating probe at 10 points in each \changedmost{cell.} Measurements ranged from 0\% - 48.1\% across the testbed. The testbed was imaged three times as the robot was teleoperated up to the edge of the boxes.


After acquiring the ground truth measurements, we masked the images and extracted the pixels contained within each box. Fig.~\ref{fig:calibration_boxes}b shows the representative spectral signature for each of the samples, plotted with $\pm$ one standard deviation from the mean box spectra. Qualitatively, the VNIR spectra show only changes in magnitude with the increasing soil moisture content. Wetter sand has a marginal albedo change. However, it is important to recall that changes in magnitude are not stable predictors, especially in partially shadowed environments. The SWIR shows more promising features, particularly the change at $\approx$ 1300 nm.

Inspired by the formulation of NDVI, we formulate an optimization problem to quantitatively choose a band pairing for a SMC index using the wavelength of the VIS-SWIR HSI system. We hypothesize that maximizing the distance between the maximum and minimum index value across bands will best allow for differentiation of intermediary moisture contents. In (\ref{eq:band_optimization}), the mathematical optimization of the band selection problem is presented. We optimize the spectral index by considering the pair of band combinations yielding the maximum difference between the dry and wet sand.

\begin{gather}
    \label{eq:band_optimization}
    f(i,j) = \biggr\rvert\biggr\rvert \frac{p_{i,wet}-p_{j,wet}}{p_{i,wet}+p_{j,wet}} - \frac{p_{i,dry}-p_{j,dry}}{p_{i,dry}+p_{j,dry}}  \biggr\rvert\biggr\rvert_{2} \\
    \underset{i,j}{\text{maximize}}\ f(i,j) \nonumber\\
    \text{subject to}\ i \in \Lambda, j \in \Lambda, i \neq j. \nonumber
\end{gather}
\begin{equation}
    \label{eq:smc}
    SMC = \frac{I_{1300} - I_{1119}}{I_{1300} + I_{1119}}
\end{equation}

$p$ is the set of pixels, with values at wavelength $i$ or $j$. The wet and dry pixels are extracted from boxes $1$ and $9$, respectively. $\Lambda$ is the set of wavelengths in the registered datacube. The optimized wavelength values align with the spectral absorption features present between $\approx 1150$ nm and $\approx 1350$ nm in the solar absorption spectra.

Extracting the optimal wavelength values yields (\ref{eq:smc}), which calculates the normalized difference between the two wavelength channels. The value of SMC here is dimensionless. In order to convert the SMC to an actual physical quantity, we consider the SMC index as the domain of a linear regression function. Fig.~\ref{fig:calibration_boxes}d depicts the results of this fit model.

From Fig.~\ref{fig:calibration_boxes}, the model shows a strong ability to distinguish the dry box from the saturated box. Predictions of the intermediary values are distinguishable, but not to the same degree as the extreme situations of dry and saturated. The distribution of the spectral values in each of these boxes shows an overlapping range of values. Given the results, it may be worth considering soil moisture as a three-class problem: dry, moist, and saturated. Considering these classes, vehicles could tailor their risk tolerances depending on which moisture levels constitute go or no-go decisions through a physically informed terramechanics model.

One of the foremost areas of future work is the extension of these methods to other soil types. The results indicate a strong ability to discern surface properties beyond broad semantic labels such as \textit{sand} or \textit{soil}. These results are produced on the assumption that the terrain class is already known. With this knowledge, SMC could then be applied to specific regions where the moisture likely affects terramechanics such as sand, soil, and clay. The prior knowledge of terrain class also reduces the computational load on the characterization algorithm, as only certain regions will need this property calculated. Additionally, per-class property modeling allows for different band combinations to be optimized for different soil types. While $1300$ and $1119$ nm are optimal for sand, the presence of organic material in other surface types might necessitate the integration of bands from the VNIR domain.

Although these example classification processes use two bands each, they leverage different parts of the spectrum, and show how the same calibrated spectral data can be utilized for both terrain identification and parameterization. Calculating these spectral indices takes $30$ ms on a CPU per datacube. This speed (30 frames per second) supports real-time processing. This ability to derive a real, physical quantity from the environment with minimal labels, minimum compute requirements, and simple modeling strongly motivates further investigation into \changedmost{HSI} in field robotics.

\section{Conclusions}
We presented a system architecture for acquiring and calibrating hyperspectral images from a moving robotic platform. The proposed method uses spectrometer–hyperspectral joint calibration to mitigate the effects of varying illumination \changedmost{and enable consistent reflectance measurements without  calibration targets.} Using this calibration, we demonstrated the estimation of \changedmost{two} terrain-relevant properties: vegetative health and soil moisture content.

This work provides a practical and computationally efficient framework for integrating hyperspectral sensing into field robotics. Future efforts will focus on extending these methods to a broader range of terrain types and developing autonomous strategies for identifying and responding to environmental conditions that affect traversability.

\bibliographystyle{IEEEtran} 
\bibliography{references}

\addtolength{\textheight}{-12cm}   

\end{document}